\documentclass[letterpaper, 10 pt, conference]{ieeeconf}  
\usepackage{colortbl} 
\usepackage{xcolor}
\usepackage{tabularx}
\usepackage{ragged2e} 
\usepackage{multirow}
\usepackage{makecell}
\usepackage{booktabs}
\usepackage[utf8]{inputenc} 
\usepackage{fontenc}
\usepackage{graphicx}  
\usepackage{subcaption} 
\usepackage{caption}  
\usepackage{float}   
\usepackage{lmodern} 
\usepackage{algorithm}
\usepackage{algpseudocode}
\usepackage{amsmath}
\usepackage{amssymb}  
\usepackage{ragged2e}
\usepackage{cite}   
\makeatletter
\let\NAT@parse\undefined
\makeatother
\usepackage[colorlinks,linkcolor=blue,citecolor=blue]{hyperref}

\justifying 
\graphicspath{{./figures/}}

\IEEEoverridecommandlockouts                              

\title{\LARGE \bf
	MemoAct: Atkinson--Shiffrin-Inspired Hierarchical Memory-Augmented Policy for Robotic Manipulation
}
\author{
	Liufan Tan, Jiale Li, and Gangshan Jing
}

\begin{document}

	\maketitle
	\begin{figure*}[htbp] 
		\centering  
		\includegraphics[width=\textwidth]{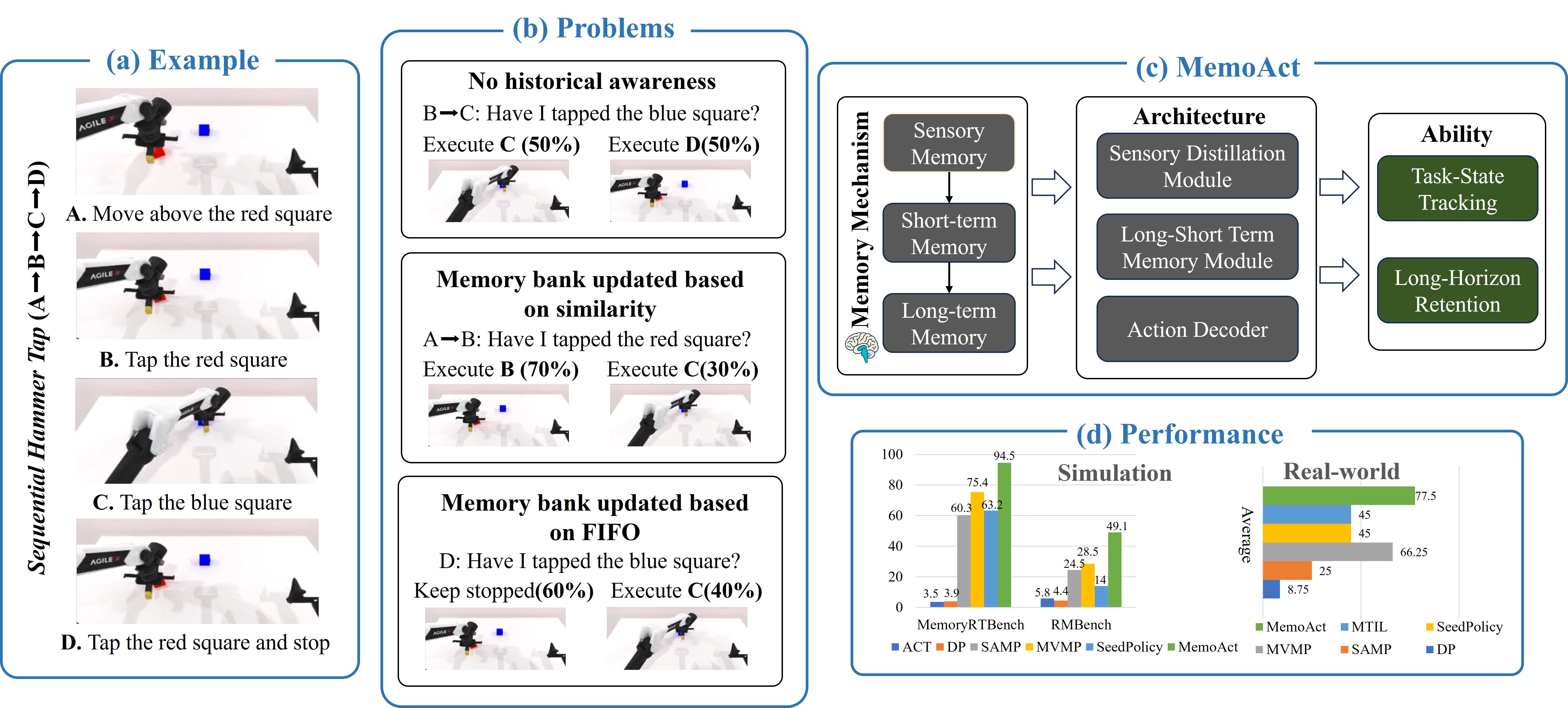}  
		
		\caption{(a) An example of a memory-dependent task. (b) Policies lacking historical awareness fail under identical observations, while existing representative memory mechanisms suffer from limited long-horizon retention and poor task-state tracking. (c) Inspired by the Atkinson--Shiffrin memory model, we propose MemoAct, which simultaneously enables precise task-state tracking and robust long-horizon retention. (d) Results on MemoryRTBench, RMBench, and real-world experiments demonstrate that MemoAct significantly outperforms baseline algorithms.}
		\label{fig:intro}  
	\end{figure*}
	

	\begin{abstract}

        Memory-augmented robotic policies are essential in handling memory-dependent tasks. However, existing approaches typically rely on simply extending the observation window, struggling to simultaneously achieve precise task-state tracking and robust long-horizon retention. To overcome these challenges, inspired by the Atkinson--Shiffrin memory model, we propose MemoAct, a hierarchical memory-augmented policy that leverages distinct memory tiers to tackle specific bottlenecks. Specifically, sensory memory filters immediate perceptual inputs, lossless short-term memory supports precise task-state tracking, and compressed long-term memory facilitates robust long-horizon retention. To enrich the evaluation landscape, we construct MemoryRTBench based on RoboTwin 2.0, comprising 6 manipulation tasks that systematically evaluate policy memory capabilities across three dimensions: sequential, spatial, and episodic memory. Extensive experiments across simulated and real-world scenarios demonstrate that MemoAct achieves superior performance compared to both existing Markovian baselines and history-aware policies. The project page is available at \url{https://tlf-tlf.github.io/MemoActPage/}.
	\end{abstract}
	
	
	\section{INTRODUCTION}	
	In recent years, robotic manipulation policies have made remarkable progress~\cite{pi0.6, DP, act, acot, lohovla, mem, song2025survey, firoozi2025foundation, ai2025review, large}, enabling robots to perform complex tasks such as folding clothes, wiping tables, and making tea. 
	These methods follow the Markov decision paradigm, predicting future actions $A_{t:t+m-1}$ solely from the current observation $O_t$.
	However, this paradigm struggles with memory-dependent tasks~\cite{mtil, memoryvla, robomme}, especially those requiring precise task-state tracking and robust long-horizon retention.
	
	Consider the ``Sequential Hammer Tap'' task in Fig.~\ref{fig:intro} (a). 
	The robot must execute four sub-tasks: (A) move the hammer above the red square, (B) tap the red square, (C) tap the blue square, and (D) tap the red square again and stop. 
	The key challenge is that the observations when tapping the red square are visually identical before and after tapping the blue square. 
	Markovian policies are prone to failure under such perceptual aliasing~\cite{history, mtil, hamlet}, as shown in Fig.~\ref{fig:intro} (b).
	
	A straightforward solution is to expand the observation horizon with a First-In-First-Out (FIFO) memory bank~\cite{sam2act, hamlet, memoryvla, cronusvla}, which provides limited historical perception for tracking recently completed sub-tasks. 
	However, as new observations overwrite old ones, information beyond the window becomes inaccessible~\cite{mtil, history}. 
	Thus, in sub-task D, the robot may forget that the blue block has already been tapped and erroneously repeat sub-task C, as shown in Fig.~\ref{fig:intro} (b). 
	
	Alternatively, MemoryVLA~\cite{memoryvla} extends the memory span by merging the most similar adjacent embeddings. 
	Although this improves long-horizon retention, such compression may discard fine-grained cues essential for task-state tracking, causing the robot to skip sub-task B and prematurely execute sub-task C, as shown in Fig.~\ref{fig:intro} (b).
	These observations motivate a natural question: can a policy combine lossless short-term memory with compressed long-term memory to achieve both precise task-state tracking and robust long-horizon retention?

	To answer this question, we draw inspiration from cognitive science, where the Atkinson--Shiffrin model organizes human memory into three tiers: sensory memory, short-term memory, and long-term memory~\cite{humanmemory, episodicmemory, workingmemory}. 
	Each tier serves a distinct role: sensory memory filters immediate perceptual inputs, short-term memory then maintains recent salient information, and long-term memory finally stores compressed experiences for persistent retention. 
	This principle is well aligned with memory-dependent manipulation, where a robot must perceive the current scene, track recent task progress, and retain distant but task-relevant history.
	
	Following this three-tier memory design, role separation and its sensory-to-short-term-to-long-term information flow, we propose MemoAct, as illustrated in Fig.~\ref{fig:intro} (c). 
	MemoAct consists of three modules: 
	(1) a sensory distillation module that extracts compact sensory memory from redundant visual inputs using a learnable query embedding;  
	(2) a long short-term memory module that maintains a lossless short-term memory bank for recent task-state tracking and a compressed long-term memory bank for long-horizon retention; 
	and (3) an action decoder that generates history-aware actions conditioned on memory-augmented embeddings.
	
	To evaluate MemoAct, we establish Memory\textbf{RT}Bench, a \textbf{R}obo\textbf{T}win 2.0-derived benchmark~\cite{robotwin2.0} with six manipulation tasks assessing sequential, spatial, and episodic memory. These three memory dimensions evaluate the ability to execute sub-tasks in the prescribed order, recall the initial scene state after intermediate operations, and repeat a specified task the required number of times, respectively.
	Experiments on MemoryRTBench, RMBench~\cite{rmbench}, and real-world scenarios show that MemoAct consistently outperforms Markovian and history-aware baselines and demonstrates reasonable generalization ability.
	Ablations verify the contributions of short- and long-term memory, and adaptation to point cloud-based policies demonstrates the plug-and-play versatility of our memory module.
	
	In summary, our contributions are threefold:
	
	1) We propose MemoAct, an Atkinson–Shiffrin-inspired visuomotor policy using hierarchical memory to overcome the limitations of simple observation window extensions in memory-dependent tasks.
	
	2) We propose a hierarchical memory mechanism that combines a fixed-size lossless short-term bank for precise task-state tracking with a compressed long-term bank for robust long-horizon retention via causal attention-based compression and similarity-based merging.
	
	3) We establish MemoryRTBench, a specialized benchmark built upon RoboTwin 2.0, comprising six manipulation tasks that systematically evaluate policy memory capabilities across sequential, spatial, and episodic dimensions. Extensive simulation and real-world experiments further demonstrate MemoAct's strong performance and plug-and-play adaptability.
	
    \begin{figure*}[htbp] 
		\centering  
		\includegraphics[width=\textwidth]{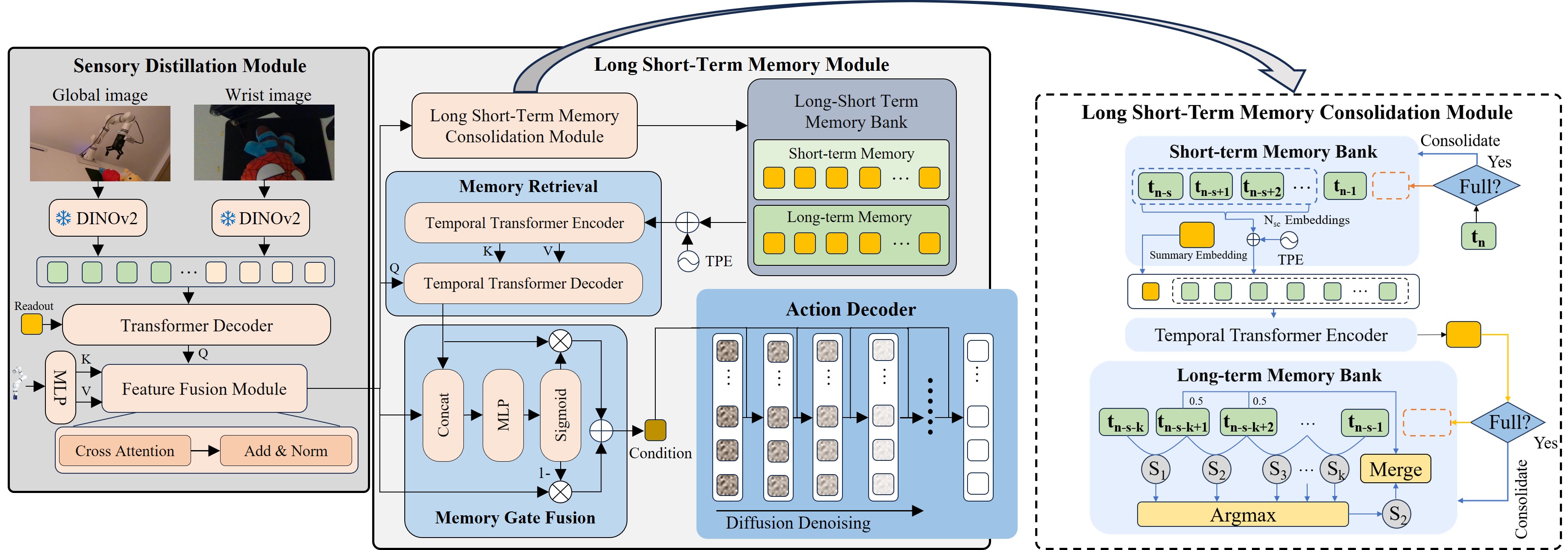}  
		
		\caption{\textbf{Overview of MemoAct Pipeline.} 
			First, the sensory distillation module encodes RGB images and proprioceptive states into high-fidelity sensory memory. 
			Subsequently, the distilled sensory memory queries relevant historical context from the long short-term memory bank. 
			Next, a gating network adaptively fuses the retrieved history with the current sensory memory to produce a condition embedding, which guides the action decoder to iteratively denoise noisy action trajectories into history-aware action trajectories. 
			Finally, the long short-term memory consolidation module updates the memory bank after each forward pass.}
		\label{fig:overview}  
	\end{figure*}
    
	\section{Related works}
	\subsection{Memory-Augmented Robotic Manipulation Policies}
	Historical perception is crucial for robotic manipulation when the current observation is insufficient for action prediction. Several approaches~\cite{sam2act, cronusvla, hamlet} employ a fixed-size FIFO memory bank. While effective for short-term dependencies, FIFO memory overwrites earlier information and thus struggles with long-horizon recall.
	To extend the temporal span, MemoryVLA~\cite{memoryvla} merges adjacent similar frames instead of using strict FIFO updates. 
	However, such merging may discard subtle cues needed for precise task-state tracking. MTIL~\cite{mtil} uses Mamba2~\cite{mamba2} to encode the full history, while SeedPolicy~\cite{seedpolicy} maintains a self-evolving latent state with gated attention to capture long-term dependencies. 
	However, compressing the entire history into fixed-size hidden states may lose fine-grained or distant task-critical information, limiting precise historical recall~\cite{wen2024rnns}.
	In contrast, MemoAct balances temporal precision and memory capacity by combining a lossless short-term memory bank for task-state tracking with a compressed long-term memory bank for long-horizon recall.
	
	\subsection{Memory-Related Robotic Manipulation Benchmarks}
	Recent efforts introduce non-Markovian manipulation benchmarks~\cite{sam2act,mikasa,rmbench,robomme}, but remain limited in different aspects. 
	MemoryBench~\cite{sam2act} has a relatively small task scale. MIKASA~\cite{mikasa} mainly targets reinforcement learning settings. RMBench~\cite{rmbench} considers different memory complexities but lacks an explicit organization by memory type, limiting fine-grained diagnosis. RoboMME~\cite{robomme} evaluates memory from multiple dimensions, but it mainly uses success rate as the evaluation metric, which can conflate memory failures with low-level control failures.
	In contrast, MemoryRTBench organizes memory requirements into three categories: sequential, spatial, and episodic memory, with six representative tasks. 
	Moreover, beyond overall success rate, we introduce memory-specific failure rates together with the low-level control error rate. This evaluation protocol allows us to separately analyze low-level control ability, task-state tracking, spatial recall, and long-horizon historical retention.
	
	\color{black}
	\section{METHOD}
	In this work, we propose MemoAct, which can be decomposed into three functions: a sensory distillation module $f_{sdm}^{\phi}$ : $O_t \rightarrow F_{O}^t$ , a long short-term memory module $f_{lstm}^{\psi}$ : $F_{O}^t, F_{M}^t \rightarrow F_C^t$, and a memory-augmented action decoder $f_{ad}^{\omega}$ : $F_{C}^t \rightarrow A_{t:t+m-1}$. The overall framework is illustrated in Fig. \ref{fig:overview}.
	
	\subsection{Sensory Distillation Module for Extracting Sensory Memory}
	Given an RGB image $I^{t} \in \mathbb{R}^{H \times W \times 3}$, we encode it with a frozen DINOv2 backbone~\cite{dinov2} into patch-level features $F_{I}^t \in \mathbb{R}^{P \times P \times C}$, following common practice~\cite{mtil, dexgraspvla}. 
	To reduce redundancy and memory cost, a Transformer decoder layer uses a single learnable readout embedding to query $F_I^t$ and compresses it into $F_{R}^t \in \mathbb{R}^{1 \times C}$. 
	Meanwhile, the robot proprioceptive state $S_t$ is projected by a MLP into $F_{S}^t \in \mathbb{R}^{1 \times C}$, aligning it with the visual embedding. 
	We then fuse the visual and proprioceptive features through an attention-style residual fusion module:
	{\small
		\begin{align}
			F_{O}^t &= \text{LN}\left( \text{Softmax}\left( \frac{(F_{R}^t W_Q) \cdot (F_{S}^t W_K)^\top}{\sqrt{d_C}} \right) \cdot (F_{S}^t W_V) + F_{R}^t \right),
			\label{eq:ffm}
		\end{align}
	}
	where $W_Q$, $W_K$, and $W_V$ are learnable projection matrices, and $d_C$ denotes the projected query dimension. 
	The fused feature $F_O^t \in \mathbb{R}^{1 \times C}$ serves as the sensory memory.

	\subsection{Long Short-Term Memory Module for Temporal Modeling}
	\textbf{Long Short-Term Memory Bank.} Formally, we define the short-term memory bank (STMB) as $M_S \in \mathbb{R}^{T_s \times C}$, with a maximum capacity of $T_s = 6$. Correspondingly, the long-term memory bank (LTMB) is denoted as $M_L \in \mathbb{R}^{T_k \times C}$, constrained by a maximum length of $T_k = 8$. We denote the collection of all embeddings stored in both the STMB and LTMB as $F_{M}^t$.
	
	\textbf{Memory Consolidation.}
	We propose a joint consolidation module to coordinate the interaction between the STMB and the LTMB, as illustrated in Fig.~\ref{fig:overview}.
	When a new observation arrives at time step $t$, it is first inserted into the STMB. Once the STMB reaches its capacity, the oldest $N_{sc}=3$ embeddings are selected for consolidation:
	{\small
		\begin{align}
			S_{sc}^t
			&=
			\left[
			s_1,
			s_2,
			\ldots,
			s_{N_{sc}}
			\right],
			\quad
			S_{sc}^t \subset M_S^t .
			\label{eq:stmb_selection}
		\end{align}
	}
	The selected embeddings are augmented with learnable temporal positional embeddings (TPE) and summarized by a two-layer causal Transformer encoder with a learnable summary embedding:
	{\small
		\begin{align}
			z_L^t
			&=
			\mathrm{TransEnc}
			\left(
			\mathrm{concat}
			\left[
			z_{\mathrm{sum}},
			S_{sc}^t + E_{\mathrm{pos}}
			\right]
			\right)[0],
			\label{eq:stmb_to_ltmb}
		\end{align}
	}
	where $E_{\mathrm{pos}}$ denotes the temporal positional embeddings, $z_{\mathrm{sum}}$ is the learnable summary embedding, and $z_L^t$ is the summary representation transferred to the LTMB. The original $N_{sc}$ embeddings are then removed from the STMB to free up capacity.
	
	When the LTMB reaches its capacity limit, we depart from conventional FIFO eviction~\cite{sam2act, hamlet}. Instead, inspired by MemoryVLA~\cite{memoryvla}, we compute the similarity between adjacent embeddings in the long-term memory bank and merge the most similar adjacent pair. Let the LTMB at time step $t$ be denoted as $M_L^t = [l_1, l_2, \ldots, l_{L}]$. The merging process is formulated as:
	{\small
		\begin{align}
			l_{i_{\mathrm{merge}}}
			&=
			\frac{1}{2}
			\left(
			l_{i_{\mathrm{merge}}}
			+
			l_{i_{\mathrm{merge}}+1}
			\right),
			\nonumber
			\\
			i_{\mathrm{merge}}
			&=
			\mathop{\arg\max}_{i=1,\ldots,L-1}
			\mathrm{cos}
			\left(
			l_i,
			l_{i+1}
			\right),
			\quad
			l_i, l_{i+1} \in M_L^t.
			\label{eq:ltmb_merge}
		\end{align}
	}
	After merging, the redundant embedding $l_{i_{\mathrm{merge}}+1}$ is removed from $M_L^t$.

	\textbf{Memory Retrieval and Gate Fusion.}
	To capture temporal dynamics in manipulation tasks, we employ a Transformer encoder-decoder architecture with causal attention.
	As illustrated in Fig.~\ref{fig:overview}, the memory context $F_M^t$ is first augmented with learnable positional embeddings and encoded by a three-layer temporal Transformer encoder:
	{\small
		\begin{align}
			H_M^t
			&=
			\mathrm{TransEnc}
			\left(
			F_M^t + E_{\mathrm{pos}}
			\right),
			\label{eq:memory_encoding}
		\end{align}
	}
	where $H_M^t$ denotes the encoded historical memory.
	
	Given the current sensory memory $F_O^t$ as the decoder query and $H_M^t$ as the encoder memory, a two-layer Transformer decoder retrieves task-relevant historical information:
	{\small
		\begin{align}
			F_{OR}^t
			&=
			\mathrm{TransDec}
			\left(
			F_O^t,
			H_M^t
			\right),
			\label{eq:memory_retrieval}
		\end{align}
	}
	where $F_{OR}^t$ is the retrieved memory feature.
	
	To adaptively fuse current perception and retrieved history, we introduce a gate fusion module:
	{\small
		\begin{align}
			\alpha^t
			&=
			\mathrm{Sigmoid}
			\left(
			\mathrm{MLP}
			\left(
			\mathrm{concat}
			\left(
			F_O^t,
			F_{OR}^t
			\right)
			\right)
			\right),
			\nonumber
			\\
			F_C^t
			&=
			\alpha^t F_{OR}^t
			+
			\left(
			\mathbf{1} - \alpha^t
			\right)
			F_O^t,
			\label{eq:gate_fusion}
		\end{align}
	}
	where $\alpha^t$ denotes the learned fusion weight.

	\subsection{Conditional Diffusion-Based Action Decoder for Action Generation}
	We formulate the action generation process as a conditional denoising procedure. Specifically, the action decoder progressively eliminates noise sampled from a Gaussian distribution $\mathcal{N}(0, \sigma^2 I)$ to reconstruct the clean action. Conditioned on the memory-augmented embedding $F_{C}^t$, the denoising update at step $k$ is defined as:
	\begin{equation}
		\begin{split}
			A_{t:t+m-1}^{k-1} 
			&= \alpha \left( A_{t:t+m-1}^{k} 
			- \gamma \epsilon_\theta \left( F_C^t, A_{t:t+m-1}^{k}, k \right) \right. \\
			&\quad \left. + \mathcal{N}(0, \sigma^2 I) \right),
		\end{split}
	\end{equation}
	where $\epsilon_\theta$ denotes a UNet-based \cite{unet} noise predictor, while $\alpha$ and $\gamma$ serve as hyperparameters governing the diffusion schedule. In this work, we employ DDPM \cite{ddpm} as our noise scheduler with 100 sampling steps. Accordingly, the training objective minimizes the Mean Squared Error (MSE) between the predicted noise and the ground-truth noise $\epsilon_k$:
	\begin{equation}
		L_{\text{diff}} = \operatorname{MSE}\left(
		\epsilon_\theta\left( F_C^t, A_{t:t+m-1}^{k}, k \right),
		\epsilon_{t:t+m-1}^{k}
		\right).
		\label{eq:mse}
	\end{equation}

	\subsection{Training}
	Diverging from the conventional practice of randomly sampling independent observation-action pairs $\{(O_t, A_{t:t+m-1})\}$, we adopt a streaming training paradigm to preserve temporal continuity. 
	In this framework, each training batch is constructed exclusively from a single episode, with samples loaded in strict chronological order rather than being randomly shuffled. 
	This ensures that the long short-term memory bank is populated sequentially, creating an ordered temporal context that is essential for effective model training. 
	To address boundary conditions where the prediction horizon $t + m - 1$ exceeds the trajectory length, we pad the sequence by replicating the terminal time step.  The long short-term memory bank is initialized as empty at the start. When it remains empty, memory retrieval and gate fusion are bypassed, with the current sensory memory $F_O^t$ directly assigned to the fused context feature $F_C^t$. We train the model with a learning rate of $1 \times 10^{-4}$.
	The detailed training procedure is outlined in Algorithm \ref{alg:MemoAct_training}.		
	
	\begin{algorithm}[htbp]
		\caption{MemoAct Training.}
		\label{alg:MemoAct_training}
		
		\textbf{Require:} Expert trajectories 
		$\mathcal{D}=\{\tau_i\}_{i=1}^{N}$, where 
		$\tau_i=\{(O_t^i,A_t^i)\}_{t=1}^{T_i}$; 
		chunk size $m$; batch size $B$; optimizer $Opt$; loss function $\mathcal{L}$; diffusion scheduler with $K$ denoising steps.\\
		\textbf{Model:} Sensory distillation module $f_{sdm}^{\phi}$; 
		long short-term memory module $f_{lstm}^{\psi}$ with memory bank $\mathcal{M}$; action decoder $f_{ad}^{\omega}$.\\
		\textbf{Return:} Trained MemoAct policy 
		$\pi_{\Phi}=\{f_{sdm}^{\phi},f_{lstm}^{\psi},f_{ad}^{\omega}\}$, 
		where $\Phi=\{\phi,\psi,\omega\}$.
		
		\begin{algorithmic}[1]
			\State Initialize policy parameters $\Phi=\{\phi,\psi,\omega\}$
			\For{each training epoch}
			\For{each trajectory $\tau_i \in \mathcal{D}$}
			\State Clear memory bank $\mathcal{M}$
			\For{$s=1$ to $T_i$ with step $B$}
			\State $e \leftarrow \min(s+B-1,T_i)$, 
			$\mathcal{L}_{batch}\leftarrow 0$
			\For{$t=s$ to $e$}
			\State $F_O^t \leftarrow f_{sdm}^{\phi}(O_t^i)$
			\State $F_C^t \leftarrow f_{lstm}^{\psi.forward}(F_O^t,\mathcal{M})$ 
			\Comment{Memory Retrieval and Gate Fusion}
			\State $Y_t \leftarrow \operatorname{Pad}(A_{t:t+m-1}^i,T_i)$
			\State Sample noise $\epsilon_t \sim \mathcal{N}(0,I)$
			\State $Y_t^k \leftarrow \operatorname{AddNoise}(Y_t,\epsilon_t,k)$
			\State $\hat{\epsilon}_t \leftarrow f_{ad}^{\omega}(F_C^t,Y_t^k,k)$
			\State $\mathcal{L}_{batch}\leftarrow 
			\mathcal{L}_{batch}+\mathcal{L}(\hat{\epsilon}_t,\epsilon_t)$
			\State $\mathcal{M}\leftarrow 
			f_{lstm}^{\psi.mc}(\mathcal{M},\operatorname{sg}(F_O^t))$ 
			\Comment{Memory Consolidation}
			\EndFor
			\State $Opt.step()$
			\EndFor
			\EndFor
			\EndFor
			\State \Return $\pi_{\Phi}$.
		\end{algorithmic}
	\end{algorithm}

	\begin{figure*}[htbp] 
		\centering  
		\includegraphics[width=\textwidth]{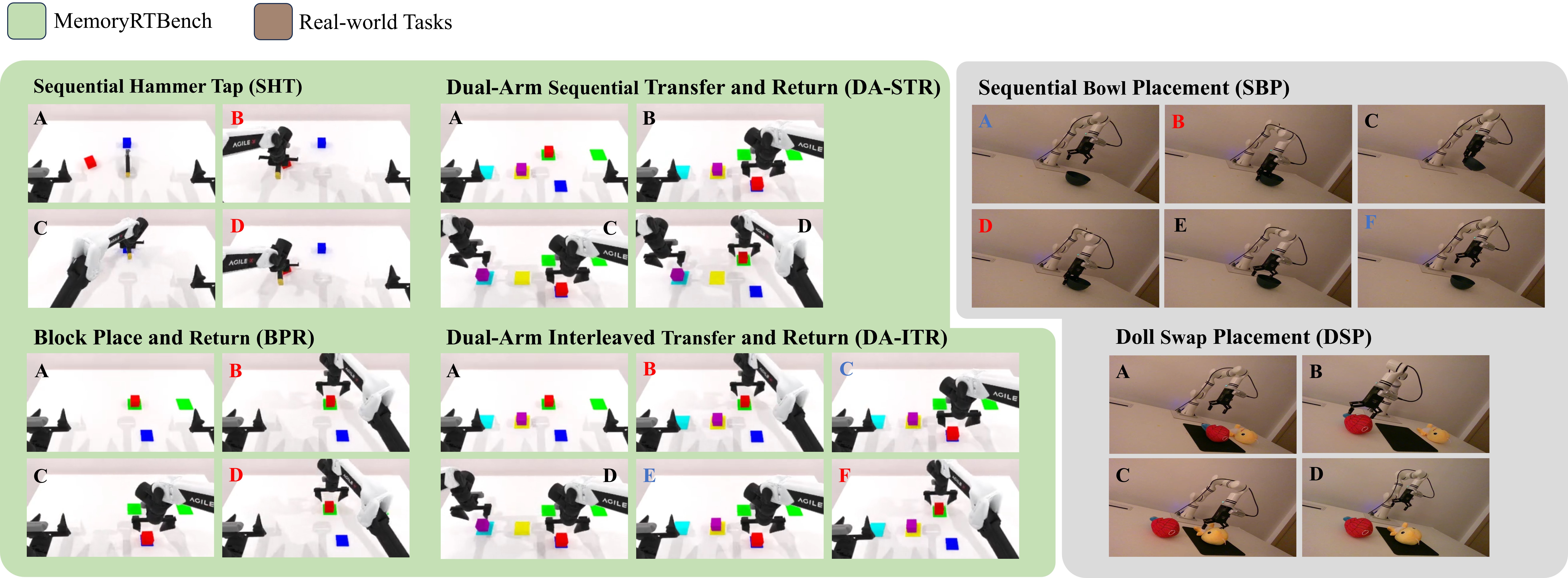}  
		
		\caption{Overview of MemoryRTBench and real-world tasks. The tasks are executed sequentially following the alphabetical order (i.e., A $\to$ B $\to$ C $\to$ ...). Notably, identical observations encountered during the execution are highlighted in red, blue or gray.}
		\label{fig:tasks}  
	\end{figure*}

	\color{black}
	\section{Experiments}
	To comprehensively evaluate the performance of MemoAct, we designed a series of experiments to address the following five core research questions:
	
	\textbf{(Q1)} How does MemoAct compare against state-of-the-art baselines on both simulated and real-world tasks?
	
	\textbf{(Q2)} How well does MemoAct generalize? Will its memory module degrade under distribution shifts?
	
	\textbf{(Q3)} How does each individual module of MemoAct contribute to the overall performance?
	
	\textbf{(Q4)} How do the capacities of long-term and short-term memory banks affect MemoAct’s performance, and do they require substantial task-dependent tuning?
	
	\textbf{(Q5)} Can the long short-term memory module of MemoAct be seamlessly integrated into other baseline policies, and does it consistently yield performance improvements in history-aware tasks?
	\subsection{Experimental Setup}
	
	\textbf{Simulation Setup.} All simulations are conducted in RoboTwin 2.0~\cite{robotwin2.0} with a Cobot Magic mobile manipulator. Visual inputs are captured via an Intel RealSense L515 RGB-D camera for MemoryRTBench and a D435 RGB-D camera for RMBench.
	
	\textbf{Real-robot Setup.} Our physical setup comprises a Realman RM65-B robotic arm with a TEK CTAG2F90-C gripper. For perception, we employ a global Intel RealSense D455 RGB-D camera and a wrist-mounted Intel RealSense D435 RGB-D camera.
	
    \textbf{Tasks and Metrics.} 
	We evaluate the policies on MemoryRTBench, RMBench~\cite{rmbench}, and several real-world experiments (see Fig.~\ref{fig:tasks} for details). 

	To diagnose whether failures on MemoryRTBench arise from memory limitations or low-level control errors, we report five metrics. 
	SeqE (\textbf{seq}uential memory \textbf{e}rror rate) measures failures in executing sub-tasks in the prescribed order, such as skipping or misordering sub-tasks. 
	SpaE (\textbf{spa}tial memory \textbf{e}rror rate) measures failures in recalling the initial scene state after intermediate operations. 
	EpiE (\textbf{epi}sodic memory \textbf{e}rror rate) measures failures in repeating a specified task the required number of times. 
	LlcE (\textbf{l}ow-\textbf{l}evel \textbf{c}ontrol \textbf{e}rror rate) measures failures caused by visual localization inaccuracies or imprecise object interactions. 
	SR (\textbf{s}uccess \textbf{r}ate) measures the probability of successfully completing all sub-tasks. Success and failure are automatically determined by predefined task-specific rules, while failure types are manually annotated from recorded execution videos.
	
	For each simulated experiment, we conduct three runs using random seeds 0, 1, and 2. 
	For each seed, we execute 50 trials per task on MemoryRTBench and 100 trials per task on RMBench. 
	For real-world scenarios, we execute 20 trials per task. All experiments are conducted on a single RTX 4090 GPU.
	
	\color{black}

	\textbf{Baselines.} We compare our method with several representative baselines, including two widely used policies, ACT~\cite{act} and DP~\cite{DP}, as well as two recent memory-enhanced policies, SeedPolicy~\cite{seedpolicy} and MTIL~\cite{mtil}. For a controlled comparison using the same policy backbone, we further build 
	MVMP (\textbf{M}emory\textbf{V}LA \textbf{M}emory-enhanced \textbf{P}olicy), which uses a similarity-based memory bank, 
	SAMP (\textbf{S}AM2\textbf{A}ct \textbf{M}emory-enhanced \textbf{P}olicy), which uses a FIFO memory bank, and a Transformer-style variant, which replaces our hierarchical memory module with a fixed-window Transformer module.

	\begin{table*}[htbp]
		\centering
		\caption{Performance comparison of MemoAct and baseline methods on MemoryRTBench across three different seeds (mean$\pm$std, \%). We only report success rates for memory-free baselines.}
		\label{tab:exp_compar_memoryrtbench_results}
		\setlength{\tabcolsep}{3pt}
		\resizebox{\linewidth}{!}{
			\begin{tabular}{l *{9}{c}}
				\toprule
				\multirow{2}{*}{\textbf{Method}}
				& \multicolumn{3}{c}{\textbf{SHT}} 
				& \multicolumn{3}{c}{\textbf{BPR}} 
				& \multicolumn{3}{c}{\textbf{STR}} \\
				\cmidrule[0.4pt](lr){2-4} 
				\cmidrule[0.4pt](lr){5-7} 
				\cmidrule[0.4pt](lr){8-10}
				& \textbf{SeqE} & \textbf{LlcE} & \textbf{SR}
				& \textbf{SpaE} & \textbf{LlcE} & \textbf{SR}
				& \textbf{SpaE} & \textbf{LlcE} & \textbf{SR} \\
				\midrule[0.4pt]
				
				\multicolumn{10}{l}{\textbf{(a) Baselines}} \\
				\midrule[0.4pt]
				
				\textbf{ACT}~\cite{act}
				& - & - & 17.3\textpm5.7
				& - & - & 2.3\textpm2.1
				& - & - & 1.0\textpm1.0 \\
				
				\textbf{DP}~\cite{DP}
				& - & - & 13.7\textpm4.2
				& - & - & 1.7\textpm1.5
				& - & - & 0.0\textpm0.0 \\
				
				\textbf{SeedPolicy}~\cite{seedpolicy}
				& 30.7\textpm5.7 & 10.0\textpm2.0 & 59.3\textpm4.5
				& 52.0\textpm5.6 & 6.3\textpm1.5 & 41.7\textpm6.8
				& 45.3\textpm3.5 & \cellcolor[gray]{0.8}\textbf{0.0\textpm0.0} & 54.7\textpm3.5 \\
				
				\midrule[0.4pt]
				\multicolumn{10}{l}{\textbf{(b) Variants of the memory module}} \\
				\midrule[0.4pt]
				
				\textbf{w/ SAM2Act-style(SAMP)}~\cite{sam2act}
				& 18.0\textpm6.6 & 6.7\textpm1.5 & 75.3\textpm6.4
				& 38.0\textpm7.5 & \cellcolor[gray]{0.8}\textbf{0.0\textpm0.0} & 62.0\textpm7.5
				& 59.0\textpm4.6 & \cellcolor[gray]{0.8}\textbf{0.0\textpm0.0} & 41.0\textpm4.6 \\
				
				\textbf{w/ MemoryVLA-style(MVMP)}~\cite{memoryvla}
				& 52.7\textpm4.2 & \cellcolor[gray]{0.8}\textbf{2.0\textpm1.7} & 45.3\textpm5.5
				& 5.0\textpm2.6 & \cellcolor[gray]{0.8}\textbf{0.0\textpm0.0} & 95.0\textpm2.6
				& 40.7\textpm6.0 & 4.3\textpm2.1 & 55.0\textpm7.5 \\
				
				\textbf{w/ Transformer-style}~\cite{transformer}
				& 5.3\textpm2.5 & 8.7\textpm1.5 & \cellcolor[gray]{0.8}\textbf{86.0\textpm2.6}
				& 41.3\textpm4.2 & \cellcolor[gray]{0.8}\textbf{0.0\textpm0.0} & 58.7\textpm4.2
				& 45.3\textpm5.1 & \cellcolor[gray]{0.8}\textbf{0.0\textpm0.0} & 54.7\textpm5.1 \\
				
				\midrule[0.4pt]
				
				\textbf{MemoAct (Ours)}
				& \cellcolor[gray]{0.8}\textbf{0.0\textpm0.0} & 15.3\textpm4.2 & 84.7\textpm4.2
				& \cellcolor[gray]{0.8}\textbf{0.0\textpm0.0} & \cellcolor[gray]{0.8}\textbf{0.0\textpm0.0} & \cellcolor[gray]{0.8}\textbf{100.0\textpm0.0}
				& \cellcolor[gray]{0.8}\textbf{0.0\textpm0.0} & \cellcolor[gray]{0.8}\textbf{0.0\textpm0.0} & \cellcolor[gray]{0.8}\textbf{100.0\textpm0.0} \\
				
				\midrule[0.4pt]
			\end{tabular}
		}
		
		\resizebox{\linewidth}{!}{
			\begin{tabular}{l *{9}{c}}
				\midrule[0.4pt]
				\multirow{2}{*}{\textbf{Method}}
				& \multicolumn{3}{c}{\textbf{ITR}} 
				& \multicolumn{3}{c}{\textbf{LBT}} 
				& \multicolumn{3}{c}{\textbf{CBTCO}} \\
				\cmidrule[0.4pt](lr){2-4} 
				\cmidrule[0.4pt](lr){5-7} 
				\cmidrule[0.4pt](lr){8-10}
				& \textbf{SpaE} & \textbf{LlcE} & \textbf{SR}
				& \textbf{EpiE} & \textbf{LlcE} & \textbf{SR}
				& \textbf{EpiE} & \textbf{LlcE} & \textbf{SR} \\
				\midrule[0.4pt]
				
				\multicolumn{10}{l}{\textbf{(a) Baselines}} \\
				\midrule[0.4pt]
				
				\textbf{ACT}~\cite{act}
				& - & - & 1.0\textpm1.0
				& - & - & 1.7\textpm1.5
				& - & - & 0.0\textpm0.0 \\
				
				\textbf{DP}~\cite{DP}
				& - & - & 2.7\textpm2.5
				& - & - & 1.0\textpm1.0
				& - & - & 1.7\textpm2.1 \\
				
				\textbf{SeedPolicy}~\cite{seedpolicy}
				& 57.7\textpm5.7 & 2.3\textpm1.5 & 40.0\textpm6.0
				& 4.0\textpm2.6 & \cellcolor[gray]{0.8}\textbf{4.0\textpm1.0} & \cellcolor[gray]{0.8}\textbf{92.0\textpm3.0}
				& 5.0\textpm2.6 & 3.7\textpm1.5 & 91.3\textpm3.5 \\
				
				\midrule[0.4pt]
				\multicolumn{10}{l}{\textbf{(b) Variants of the memory module}} \\
				\midrule[0.4pt]
				
				\textbf{w/ SAM2Act-style(SAMP)}~\cite{sam2act}
				& 40.0\textpm8.2 & 12.0\textpm3.6 & 48.0\textpm10.5
				& \cellcolor[gray]{0.8}\textbf{0.0\textpm0.0} & 17.3\textpm4.2 & 82.7\textpm4.2
				& 42.7\textpm5.7 & 4.3\textpm1.5 & 53.0\textpm6.0 \\
				
				\textbf{w/ MemoryVLA-style(MVMP)}~\cite{memoryvla}
				& 9.3\textpm3.2 & \cellcolor[gray]{0.8}\textbf{0.0\textpm0.0} & 90.7\textpm3.2
				& 6.7\textpm3.1 & 8.0\textpm1.0 & 85.3\textpm3.8
				& 16.7\textpm5.7 & 2.3\textpm1.5 & 81.0\textpm6.0 \\
				
				\textbf{w/ Transformer-style}~\cite{transformer}
				& 35.3\textpm5.1 & \cellcolor[gray]{0.8}\textbf{0.0\textpm0.0} & 64.7\textpm5.1
				& \cellcolor[gray]{0.8}\textbf{0.0\textpm0.0} & 12.7\textpm4.0 & 87.3\textpm4.0
				& 46.7\textpm7.0 & \cellcolor[gray]{0.8}\textbf{0.0\textpm0.0} & 53.3\textpm7.0 \\
				
				\midrule[0.4pt]
				
				\textbf{MemoAct (Ours)}
				& \cellcolor[gray]{0.8}\textbf{0.0\textpm0.0} & 1.3\textpm2.3 & \cellcolor[gray]{0.8}\textbf{98.7\textpm2.3}
				& \cellcolor[gray]{0.8}\textbf{0.0\textpm0.0} & 9.3\textpm4.2 & 90.7\textpm4.2
				& \cellcolor[gray]{0.8}\textbf{0.0\textpm0.0} & 7.3\textpm4.2 & \cellcolor[gray]{0.8}\textbf{92.7\textpm4.2} \\
				
				\bottomrule
			\end{tabular}
		}
	\end{table*}
	
	\subsection{Results}
	
	\textbf{Superiority of MemoAct over Baselines (Q1).}
	As shown in Tables~\ref{tab:exp_compar_memoryrtbench_results}, \ref{tab:exp_compar_rmb_results}, \ref{tab:real_world_results}, and Fig.~\ref{fig:intro} (d), MemoAct achieves the best overall performance across simulation and real-world evaluations. 
	It obtains average success rates of $94.5\%$, $49.1\%$, and $77.5\%$ on MemoryRTBench, RMBench, and real-world tasks, outperforming the strongest baseline MVMP~\cite{memoryvla} by $19.1\%$, $20.6\%$, and $11.25\%$, respectively.
	
	Methods without explicit memory modeling, such as ACT~\cite{act} and DP~\cite{DP}, perform poorly, achieving only $3.9\%$/$3.5\%$ on MemoryRTBench and $5.8\%$/$4.4\%$ on RMBench. 
	This confirms the necessity of historical context for memory-dependent manipulation. 
	Recent memory-enhanced policies outperform ACT and DP but remain task-dependent.
	SeedPolicy~\cite{seedpolicy} performs well on episodic-memory tasks such as LBT ($92.0\%$) and CBTCO ($91.3\%$), but degrades on spatial-memory tasks such as BPR ($41.7\%$) and ITR ($40.0\%$).
	MTIL~\cite{mtil} achieves $70\%$ on real-world CTBO, yet its average success rate is only $45\%$, far below MemoAct's $77.5\%$.
	We attribute these limitations to their compact recurrent-style memory design, where the entire history is compressed into hidden states, leading to imprecise retention of fine-grained historical information.
	
	Among the controlled memory baselines, SAMP and the Transformer-style variant show relatively strong task-state tracking ability. 
	SAMP achieves 75.3\% success on SHT, while the Transformer-style variant obtains the best SHT success rate of 86.0\%. 
	However, both lack explicit long-horizon retention: SAMP may discard early evidence due to FIFO updates, and the Transformer-style variant is limited by fixed-window history modeling, leading to degraded performance on long-horizon tasks. This suggests that simply modeling the history with a generic transformer is insufficient: without explicit memory consolidation, the model struggles to retain fine-grained historical cues.
	MVMP shows stronger retention, achieving $95.0\%$ on BPR and $90.7\%$ on ITR, but suffers from sequential tracking errors, e.g., $52.7\%$ SeqE on SHT and only $21.0\%$ success on the RMBench task Rearrange Blocks. 
	MemoAct avoids these limitations by combining lossless short-term memory with compressed long-term memory. 
	MemoryRTBench diagnostics show that MemoAct reduces memory-related errors to 0.0\% across sequential, spatial, and episodic memory dimensions. 
	Real-world experiments further validate its effectiveness, where MemoAct achieves the best or tied-best performance on all four tasks. 
	Overall, these results demonstrate that the proposed long-short term memory design enables robust history-aware manipulation.
	
	\begin{table}[htbp]
		\centering
		\small
		\caption{Performance comparison results of MemoAct and baseline methods on the RMBench across three different seeds (mean$\pm$std, \%).}
		\label{tab:exp_compar_rmb_results}
		\resizebox{\linewidth}{!}{
			\begin{tabular}{lcccccc}
				\toprule
				\textbf{Tasks \textbackslash\ Methods} 
				& \textbf{ACT}~\cite{act} 
				& \textbf{DP}~\cite{DP} 
				& \textbf{SAMP}~\cite{sam2act} 
				& \textbf{MVMP}~\cite{memoryvla} 
				& \textbf{SeedPolicy}~\cite{seedpolicy} 
				& \textbf{MemoAct (ours)} \\
				\midrule
				
				\textbf{Put Back Block}      
				& 0.0\textpm0.0   
				& 0.0\textpm0.0   
				& 23.3\textpm4.0    
				& 34.0\textpm4.6  
				& 35.0\textpm4.0                                                 
				& \cellcolor[gray]{0.8}\textbf{41.0\textpm4.0} \\
				
				\textbf{Swap T}              
				& 2.3\textpm2.5   
				& 16.0\textpm3.6  
				& 25.3\textpm3.5    
				& \cellcolor[gray]{0.8}\textbf{55.0\textpm4.0}    
				& 12.7\textpm3.5 
				& 53.3\textpm1.5 \\
				
				\textbf{Rearrange Blocks}    
				& 19.3\textpm3.5  
				& 0.0\textpm0.0   
				& 46.7\textpm4.5    
				& 21.0\textpm3.6      
				& 6.3\textpm2.5                             
				& \cellcolor[gray]{0.8}\textbf{98.0\textpm2.0} \\
				
				\textbf{Observe and Pick Up} 
				& 1.7\textpm2.1   
				& 1.7\textpm1.5   
				& 2.7\textpm2.1     
				& \cellcolor[gray]{0.8}\textbf{4.0\textpm2.0}  
				& 2.0\textpm2.0   
				& \cellcolor[gray]{0.8}\textbf{4.0\textpm2.0} \\
				
				\midrule
				
				\textbf{Average}                      
				& 5.8\textpm2.0 
				& 4.4\textpm1.3 
				& 24.5\textpm3.5 
				& 28.5\textpm3.6  
				& 14.0\textpm3.0                              
				& \cellcolor[gray]{0.8}\textbf{49.1\textpm2.4} \\
				
				\bottomrule
			\end{tabular}
		}
	\end{table}
	
	\begin{table}[htbp]
		\centering
		\small
		\caption{Performance comparison results of MemoAct and baseline methods on real-world tasks (\%).}
		\label{tab:real_world_results}
		\resizebox{\linewidth}{!}{
			\begin{tabular}{l c c c c c c}
				\toprule
				\textbf{Tasks \textbackslash\ Methods} & \textbf{DP}~\cite{DP} & \textbf{SAMP~\cite{sam2act}} & \textbf{MVMP~\cite{memoryvla}} &
				\textbf{SeedPolicy~\cite{seedpolicy}} &
				\textbf{MTIL~\cite{mtil}} & \textbf{MemoAct (ours)} \\
				\midrule
				\textbf{CTBO} & $35$ & $10$ & $45$   & $60$   &\cellcolor[gray]{0.8}$\textbf{70}$   & \cellcolor[gray]{0.8}$\textbf{70}$ \\
				\textbf{PBB} & $0$ & $20$ & \cellcolor[gray]{0.8}$\textbf{70}$   & $15$   & $10$   & \cellcolor[gray]{0.8}$\textbf{70}$ \\
				\textbf{GRB} & $0$ & $45$ & $90$   &$85$   &$65$   & \cellcolor[gray]{0.8}$\textbf{95}$ \\
				\textbf{DSP} & $0$ & $25$ & $60$   &$20$   &$35$   & \cellcolor[gray]{0.8}$\textbf{75}$ \\
				\midrule
				\textbf{Average}      & $8.75$ & $25$& $66.25$ & $45$ &$45$  & \cellcolor[gray]{0.8}$\textbf{77.5}$ \\
				\bottomrule
			\end{tabular}
		}
	\end{table}
	
	\subsection{Generalization}
	\textbf{MemoAct demonstrates certain generalization ability, and its memory module remains effective under distribution shifts (Q2).}
	As shown in Table~\ref{tab:memoact_general_results}, MemoAct maintains comparable performance under object-color shifts in PBB, with success rates of $70\%$ for yellow and $65\%$ for green, compared to $70\%$ in the original setting.
	For CTBO, the success rate drops from $70\%$ to $55\%$ under background shift, and our failure analysis shows that this drop is mainly due to inaccurate visual localization rather than memory-related errors.
	These results suggest that moderate visual shifts do not impair MemoAct's memory capability, while the final task performance is still limited by the robustness of visual perception.

	\begin{table}[htbp]
		\centering
		\small
		\caption{Performance of MemoAct in generalization experiments (\%).}
		\label{tab:memoact_general_results}
		
		\resizebox{\linewidth}{!}{
			\begin{tabular}{lccc cc}
				\toprule[1.2pt]
				\textbf{Method \textbackslash\ Tasks} & \multicolumn{3}{c}{\textbf{PBB }} & \multicolumn{2}{c}{\textbf{CTBO}} \\
				\cmidrule(lr){2-4} \cmidrule(lr){5-6}
				& In-Domain & Red-to-Yellow & Red-to-Green & In-Domain & Background-Shift \\
				\midrule[0.5pt]
				\textbf{MemoAct} & \cellcolor[gray]{0.8}\textbf{70} & \cellcolor[gray]{0.8}\textbf{70} & 65 ($\downarrow$5) & \cellcolor[gray]{0.8}\textbf{70} & 55 ($\downarrow$15) \\
				\bottomrule[1.2pt]
			\end{tabular}
		}
	\end{table}
	
	\subsection{Ablation}
	\textbf{We further investigate the specific contributions of individual modules within MemoAct (Q3)}, as summarized in Table~\ref{tab:exp_abla_results}.
	Replacing DINOv2~\cite{dinov2} with ResNet-18~\cite{resnet} reduces the average success rate from $94.5\%$ to $75.7\%$, highlighting the importance of strong visual representations.
	Within the memory consolidation module, removing TPE or replacing the temporal transformer encoder with simple addition lowers the average success rate to $79.7\%$, with clear degradation on temporally or episodically demanding tasks such as SHT and CBTCO.
	This verifies that explicit temporal modeling is crucial for preserving task order and consolidating historical observations.
	Replacing the gating network with direct addition further reduces the average success rate to $81.4\%$, indicating that adaptive feature aggregation is necessary for selectively integrating current observations with memory.
	
	\begin{table}[htbp]
		\centering
		\small
		\caption{Ablation experimental results evaluating different variants of MemoAct across three different seeds (mean$\pm$std, \%).}
		\label{tab:exp_abla_results}
		\resizebox{\linewidth}{!}{
			\begin{tabular}{lccccccc}
				\toprule
				\textbf{Variants \textbackslash\ Tasks} 
				& \textbf{SHT} 
				& \textbf{BPR} 
				& \textbf{STR} 
				& \textbf{ITR} 
				& \textbf{LBT} 
				& \textbf{CBTCO} 
				& \textbf{Average} \\
				\midrule
				
				\textbf{DINOv2 \cite{dinov2} $\rightarrow$ ResNet18 \cite{resnet}} 
				& 79.7\textpm4.5 
				& 62.0\textpm6.6 
				& 79.3\textpm4.7 
				& 90.3\textpm3.5 
				& 83.0\textpm6.6 
				& 59.7\textpm7.0 
				& 75.7\textpm5.5 \\
				
				\textbf{w/o TPE} 
				& 77.0\textpm5.6 
				& 97.0\textpm2.6 
				& 97.7\textpm3.2 
				& 95.3\textpm4.7 
				& 82.7\textpm4.0 
				& 28.7\textpm6.0 
				& 79.7\textpm4.4 \\
				
				\textbf{Consol. TransEnc. $\rightarrow$ Add} 
				& 71.0\textpm5.6 
				& 98.7\textpm1.5 
				& 95.7\textpm4.5 
				& 97.0\textpm3.6 
				& 86.7\textpm5.0 
				& 29.0\textpm5.0 
				& 79.7\textpm4.2 \\
				
				\textbf{Gate $\rightarrow$ Add} 
				& 81.7\textpm4.2 
				& 99.0\textpm1.0 
				& 96.3\textpm3.1 
				& 97.7\textpm2.5 
				& 88.7\textpm4.5 
				& 25.0\textpm6.6 
				& 81.4\textpm3.7 \\
				
				\midrule
				
				\textbf{MemoAct (ours)} 
				& \cellcolor[gray]{0.8}\textbf{84.7\textpm4.2} 
				& \cellcolor[gray]{0.8}\textbf{100.0\textpm0.0} 
				& \cellcolor[gray]{0.8}\textbf{100.0\textpm0.0} 
				& \cellcolor[gray]{0.8}\textbf{98.7\textpm2.3} 
				& \cellcolor[gray]{0.8}\textbf{90.7\textpm4.2} 
				& \cellcolor[gray]{0.8}\textbf{92.7\textpm4.2} 
				& \cellcolor[gray]{0.8}\textbf{94.5\textpm2.5} \\
				
				\bottomrule
			\end{tabular}
		}
	\end{table}

\begin{figure}[htbp] 
		\centering  
		\includegraphics[width=.5\textwidth]{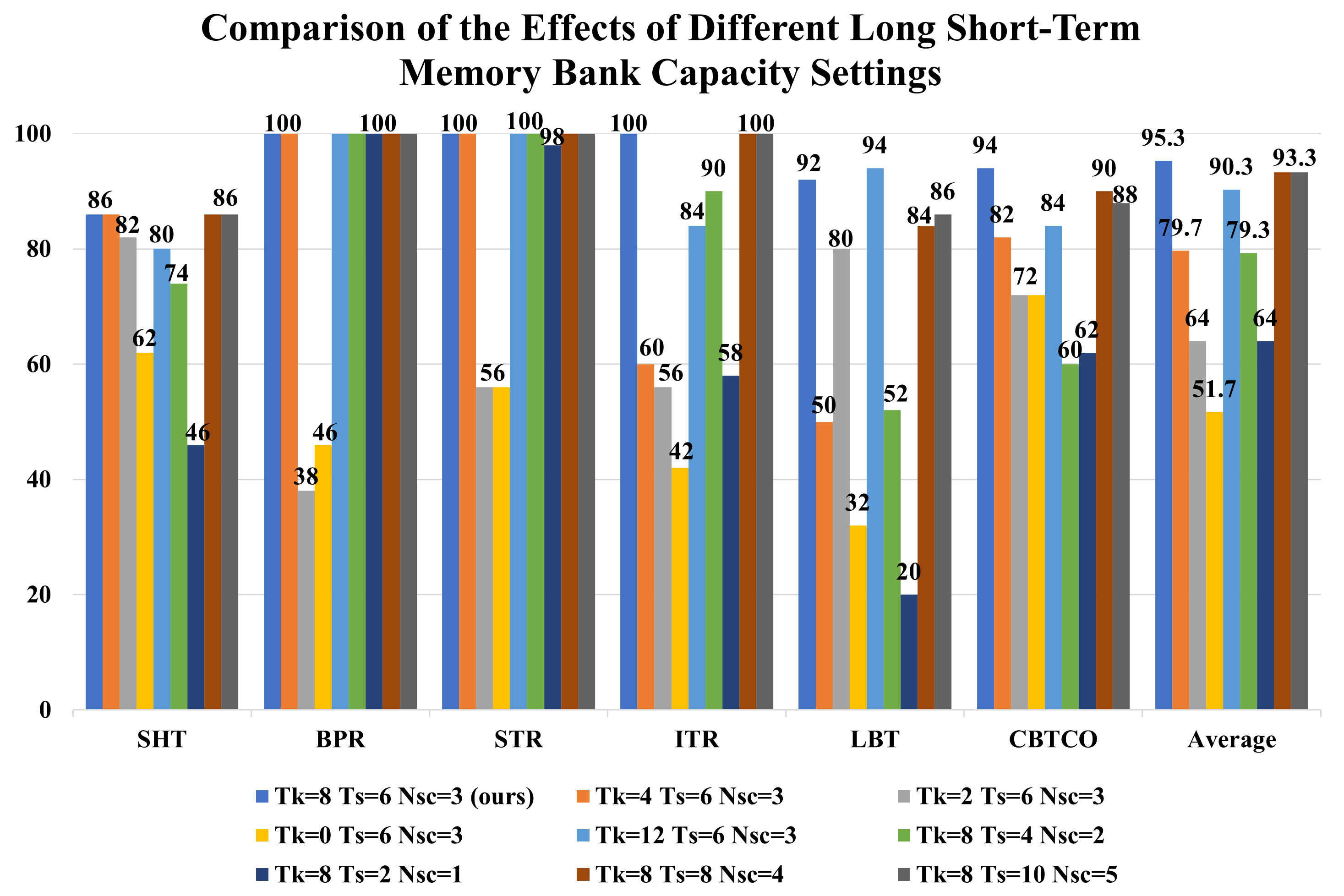}  
		
		\caption{Performance comparison of MemoAct under different memory capacities (\%), evaluated with seed 0.
		}
		\label{fig:comparsion}  
	\end{figure}

	\textbf{The long-term and short-term memory capacities control long-horizon retention and task-state tracking, respectively (Q4).}
	As shown in Fig.~\ref{fig:comparsion}, the capacity ablation shows that the long-term memory bank and short-term memory bank play different roles. 
	
	The long-term memory capacity $T_k$ mainly affects tasks that require long-horizon recall. 
	When reducing $T_k$ from $8$ to $4$ and $2$, the average success rate drops from $95.3\%$ to $79.7\%$ and $64.0\%$, respectively. 
	In particular, BPR, STR, and ITR, which require recalling early observations or initial task states, show an overall decreasing trend as $T_k$ becomes smaller. 
	In contrast, the success rate of SHT remains relatively stable with only a slight decrease when $T_k$ is reduced from $8$ to $4$ and $2$. 
	This confirms that long-term memory is especially important for retaining task-critical historical cues. 

	The short-term memory capacity $T_s$ mainly affects recent task-state tracking. 
	When reducing $T_s$ from $6$ to $4$ and $2$, the average success rate drops from $95.3\%$ to $79.3\%$ and $64.0\%$, respectively. 
	Tasks that heavily rely on task-state tracking, such as SHT, also show a substantial performance drop under smaller $T_s$ settings. 
	By contrast, BPR, STR, and ITR remain relatively stable compared with the sharp degradation observed in SHT. 
	This indicates that short-term memory plays a more direct role in maintaining recent execution context.
	
	\textbf{
	MemoAct can work effectively with a unified default configuration, while moderate task-specific adjustment may further benefit certain tasks (Q4). }
	Our default setting, $T_k=8$, $T_s=6$, and $N_{sc}=3$, achieves the best average success rate. 
	Its consistent superiority over other baselines on RMBench and real-world tasks further validates the effectiveness of this default configuration. 
	Meanwhile, settings close to the default capacity achieve comparable average performance, whereas overly small memory banks lead to larger drops. 
	This shows that MemoAct is robust within a reasonable range of memory capacities, but can be affected when the memory configuration deviates too much from a balanced setting.
	
	Nevertheless, if task-specific tuning is needed, the results provide a practical tuning guideline under a unified memory budget: tasks dominated by state-tracking errors benefit from larger $T_s$, while tasks requiring recall of early evidence benefit from larger $T_k$.

	\subsection{Extension}
	\textbf{Our memory module can be seamlessly integrated into existing policies to improve history-aware decision making (Q5).}
	We integrate our memory module into DP3~\cite{DP3}.
	As shown in Table~\ref{tab:dp3_seed_stats}, our memory module improves DP3's average success rate on MemoryRTBench from 22.1\% to 60.1\%.
	 
	\begin{table}[htbp]
		\centering
		\small
		\caption{Performance comparison between DP3 and
			DP3 + memory module (ours) on MemoryRTBench across three different seeds (mean$\pm$std, \%).}
		\label{tab:dp3_seed_stats}
		\resizebox{\linewidth}{!}{
			\begin{tabular}{lccccccc}
				\toprule
				\textbf{Methods \textbackslash\ Tasks} & \textbf{SHT} & \textbf{BPR} & \textbf{STR} & \textbf{ITR} & \textbf{LBT} & \textbf{CBTCO} & \textbf{Average} \\
				\midrule
				DP3~\cite{DP3}
				& 8.3\textpm3.5 
				& 46.7\textpm6.0 
				& 50.3\textpm4.5 
				& 2.0\textpm2.0 
				& 16.3\textpm4.5 
				& 9.0\textpm4.6 
				& 22.1\textpm1.0 \\
				\cellcolor[gray]{0.8}\textbf{DP3 + memory module (ours)} 
				& \cellcolor[gray]{0.8}\textbf{78.0\textpm7.0} 
				& \cellcolor[gray]{0.8}\textbf{96.0\textpm4.0} 
				& \cellcolor[gray]{0.8}\textbf{52.3\textpm5.5} 
				& \cellcolor[gray]{0.8}\textbf{75.3\textpm6.0} 
				& \cellcolor[gray]{0.8}\textbf{21.0\textpm5.6} 
				& \cellcolor[gray]{0.8}\textbf{37.7\textpm6.5} 
				& \cellcolor[gray]{0.8}\textbf{60.1\textpm3.6} \\
				\bottomrule
			\end{tabular}
		}
	\end{table}
	
	\section{CONCLUSIONS}
	In this work, we propose MemoAct, a memory-augmented visuomotor policy, and MemoryRTBench, a benchmark for history-aware manipulation.
	Experiments in simulation and real-world settings demonstrate MemoAct's effectiveness over existing methods.
	Future work will explore distilling historical interactions into reusable robotic experience for lifelong learning.

	



	\bibliographystyle{IEEEtran}
	\bibliography{references}

\end{document}